# Weakly-Supervised Learning via Multi-Lateral Decoder Branching for Tool Segmentation in Robot-Assisted Cardiovascular Catheterization

Olatunji Mumini Omisore, Toluwanimi Akinyemi, Anh Nguyen, Lei Wang

*Abstract*— **Robot-assisted catheterization has garnered a good attention for its potentials in treating cardiovascular diseases. However, advancing surgeon-robot collaboration still requires further research, particularly on task-specific automation. For instance, automated tool segmentation can assist surgeons in visualizing and tracking endovascular tools during procedures. While learning-based models have demonstrated state-of-the-art segmentation performances, generating ground-truth labels for fully-supervised methods is labor-intensive, time consuming, and costly. In this study, we developed a weakly-supervised learning method that is based on multi-lateral pseudo labeling for tool segmentation in cardiovascular angiogram datasets. The method utilizes a modified U-Net architecture featuring one encoder and multiple laterally branched decoders. The decoders generate diverse pseudo labels under different perturbations to augment the available partial annotation for model training. A mixed loss function with shared consistency was adapted for this purpose. The weakly-supervised model was trained end-to-end and validated using partially annotated angiogram data from three cardiovascular catheterization procedures. Validation results show that the weakly-supervised model could perform closer to fully-supervised models. Furthermore, the proposed multi-lateral approach outperforms three well known weakly-supervised learning methods, offering the highest segmentation performance across the three angiogram datasets. Numerous ablation studies confirmed the model's consistent performance under different settings. Finally, the model was applied for tool segmentation in a robot-assisted catheterization experiments. The model enhanced visualization with high connectivity indices for guidewire and catheter, and a mean segmentation time of 35.26±11.29 ms per frame. This study provides a fast, stable, and less expensive method for segmentation and visualization of endovascular tools in robot-assisted cardiac catheterization.**

## I. Introduction

As a major cause of morbidities and mortalities, cardiovascular diseases have received a significant amount of attention in the recent years [1]. To address the challenges of open surgery —the traditional treatment method, intelligent surgical robots and advanced imaging methods are being used for cardiovascular intervention. The approaches involve use of X-ray, computed tomography, or magnetic resonance imaging for endovascular catheterization and evaluation procedures [2]. Medical imaging modalities enable non-invasive visualization and inspection of the cardiac system during computer-assisted diagnosis, planning, and treatment. Throughout each stage, cardiovascular angiograms are acquired, and advanced image processing methods are essential for structural interpretation and quantification [3]. Similarly, fast and accurate image processing methods are helpful for tool visualization and tracking during interventions. Image processing tasks include registration, segmentation or reconstruction of blood vessels and endovascular tools during fluoroscopy. On segmentation, different *physics-based* and *learning-based* methods have been developed. Classically, *physics-based* methods classify cardiac structures in grayscale or RGB image frames using pixel-level intensity thresholding or region clustering. These methods often involve manual tasks (*e.g.* generating bounding box) to reduce computational complexities, making them not suitable for clinical applications with large and dynamic data.

Learning-based segmentation methods have contributed to significant advancement in cardiovascular angiogram analysis [4, 5]. With initiative of fully-supervised learning, Zhou *et al.* [6] developed the concept of pyramid attention recurrent networks for tool segmentation and tracking in x-ray images. Ronneberger *et al.* [7] proposed U-Net, an architecture that uses contracting and expanding paths for precise pixel-level segmentation and localization in medical imaging. The architecture was extended with nested dense skip paths and deep supervision for robust medical imaging applications [8]. These learning-based models are capable of capturing fine grained details of foreground pixels at low level resolution. However, the models are less sensitive to boundary preservation. To address this, Gu *et al.* [9] developed context encoder network to capture high-level information for better spatial details preservation in medical image segmentation. Yet, this network only outperformed classical U-Net in retina disc and lung segmentation [9, 10]. While the learning-based methods have significantly improved segmentation accuracy in medical imaging, creating a generalized model for dynamic imaging data with distribution mismatch and class imbalance is hard [11]. The above-mentioned networks only considered local feature contexts while performance and computational overhead issues presents as their limitations. Thus, the models are unsuitable for leveraging long range dependencies in processing cardiac angiograms. Furthermore, fully-supervised learning places a huge burden on surgeons to create high-quality masks required for model training and validation.

Weakly-supervised learning offer a potential solution to these challenges in medical image processing, especially when required for computer-assisted interventions [12-14]. By using sparse or partial annotations instead of fully annotated masks, weakly-supervised methods can reduce the time and effort required from domain experts. Current approaches include training models with point-wise, scribble-wise, and bounding box annotations [12]. Qu *et al.* [15] developed a weakly-supervised segmentation model trained with point annotations, employing a two-stage system with a self-supervised model

This work was supported by the National Natural Science Foundation of China and the Ministry of Science and Technology of China.

O. M. Omisore, T. O. Akinyemi, and L. Wang are with Research Center for Medical Robotics and Minimally Invasive Surgical Devices, Shenzhen Institute of Advanced Technology, Chinese Academy of Sciences, China.

A Nguyen is with the Smart Robotic Lab, Department of Computer Science, University of Liverpool, United Kingdom..

Corresponding Author: Olatunji Omisore; Email: omisore@siat.ac.cn.

and Gaussian masking for nuclei detection. He *et al.* [16] applied a self-teaching strategy for pixel-level segmentation in sparsely annotated 2D cardiac images. Weakly-supervised learning approaches either rely on sparsely annotated data or a combination of a small number of fully annotated signals with a large unlabeled portion to produce feature maps that can yield comparable segmentation results to fully-supervised methods. For instance, Viniavskyi *et al.* [17] developed a supervised model to generate image-level pseudo labels for abnormal chest regions in X-ray images. Subsequently, activation maps were propagated for automatic localization of lesions. The model used dual output branches to predict both displacement vector fields and the class boundaries.

Bounding-box annotations, which was widely used prior to the advent of learning-based segmentation [18, 19], have also been applied for developing generalized and robust weakly-supervised models. Typically, it involves using three or more coordinates around an object of interest in an image, and training a model to detect boundaries. Bounding-box annotations have been applied in different areas of medical imaging [13, 14, 20, 21]. However, deciding a certain region-ground separation for bounding boxes remains challenging due to lack of supervisory signals. Scribble-based methods are actively being investigated for weakly-supervised semantic segmentation in medical imaging. Luo *et al*. [12] showed how scribbles can be used for weakly-supervised learning and segmentation of cardiac images. Scribble annotations can enhance state-of-the-art (SOTA) learning architectures like U-Net, U-Net++, and DeepLabV3+ that are widely used for medical image segmentation. Additionally, scribble-based supervision signals can leverage weakly-supervised learning with global regularization, multi-scale attention and mixed augmentation consistency methods [22-24] to improve vessel and tool segmentation performance in cardiovascular images. This approach alleviates the need of dense annotation for fully-supervised learning while delivering comparable results. Alternatively, weak supervision can be achieved with point annotations in medical images. In addition to Qu *et al.* [15], Zhai *et al*. [25] employed point annotations for weakly-supervised learning in medical image processing, though their used only a few point masks to define the segmentation targets. Contextual regularization with conditional random field and variance minimization were applied for consistency learning [22]. Issam *et al.* [26] demonstrated that single pixel annotation, combined with consistency learning, can regularize weakly-supervised models and stabilize segmentation outputs. Point annotation is easier and faster compared to scribble and bounding-box methods.

Learning from sparse annotations presents significant challenges and requires effective regularization techniques. The application of existing weakly-supervised methods in cardiovascular catheterization imaging remains limited. For instance, Yang *et al*. [27] employed voxel labels generated through line filtering, which were updated iteratively to produce class activation feature maps for segmentation of catheter in frustum ultrasound images. The study relied on bounding-box annotations, which resulted in noisy masks and a model whose performance fell sizably short of SOTA fully-supervised models. Also, this approach was not demonstrated for additional flexible endovascular tools, such as guidewire, which produces highly deformable shapes during catheterization.

In this study, we propose a weakly-supervised learning method for concurrent pseudo labeling and segmentation of endovascular tools in cardiovascular angiograms. This can enhance existing backbone networks by utilizing a single encoder and multiple decoders to generate feature embedding from partial annotations for pixel-level segmentation. The decoders are perturbed to capture complementary features imposed with shared consistency regularization to create a robust and well-generalized segmentation model. A U-Net architecture with one encoder and multiple laterally branched decoders was utilized for this study. Multi-scale pixel-wise predictions are regularized through a mixed loss function that generates pseudo labels for end-to-end model training. The trained model was applied for tool segmentation in angiogram data obtained during catheterization in synthetic human aorta [28] and robot-assisted trials in rabbit and pig [29]. The key contributions of this study are as follows:

*1)* The development of a weakly-supervised method with multi-lateral branched decoders for endovascular tool segmentation in partially-annotated cardiac angiograms.
*2)* Introduction of a shared consistency term to integrate the supervision signals generated from multiple decoders.
*3)* Validation and ablation studies conducted to assess the performance and stability of the proposed method on various cardiovascular angiogram datasets, and
*4)* Deployment of the proposed weakly-supervised method, demonstrating its application for tool segmentation during robot-assisted catheterization in an aorta phantom.

The remainder of this paper is organized as follows: Section II introduces the proposed weakly-supervised multi-lateral learning method. Sections III and IV present the validation and ablation studies conducted for performance analysis of the method. Section V discusses application of the method in a robot-assisted catheterization setup. Finally, the conclusion of the study and future works are in Section VI.

## II. Proposed Method

We utilized angiogram datasets with partial annotations to leverage the expertise of domain experts. Assuming that data instances are drawn from a Gaussian mixture model $\mathcal{J}(x|\Theta)$ with $n$ classes in Eq. (1), the soft pseudo labels for each pixel $x$ can be generated and propagated to train a deep network. $\beta_i$ is taken as a mixture coefficient such that $\sum_{i=1}^{n}\beta_i = 1$, while $\Theta = \{\theta_i\}$ denotes the network parameters. To assign each pixel to a class, it is necessary to compare the labeled pixels in the partially annotated signals with respect to the unlabeled components in the angiograms.

$$\mathcal{J}(x|\Theta) = \sum_{i=1}^{n} \beta_i \mathcal{J}(x|\theta_i) \tag{1}$$

To determine the class of each pixel, we considered that the label $y_i$ of pixel $i$ as a random variable whose distribution $P(y_i|x_i, g_i)$ is determined by the mixture component $g_i$ and the pixel's features embedding. Using maximum *a posteriori criterion*, the unlabeled part of $y_i$ can be estimated based on the supervised signals, as follows:

$$h(x) = \underset{c\in[0,1]}{argmax} \sum_{i=1}^{n} P(y_i = c|g_i = j, x_i) \times P(g_i = j|x_i). \tag{2}$$

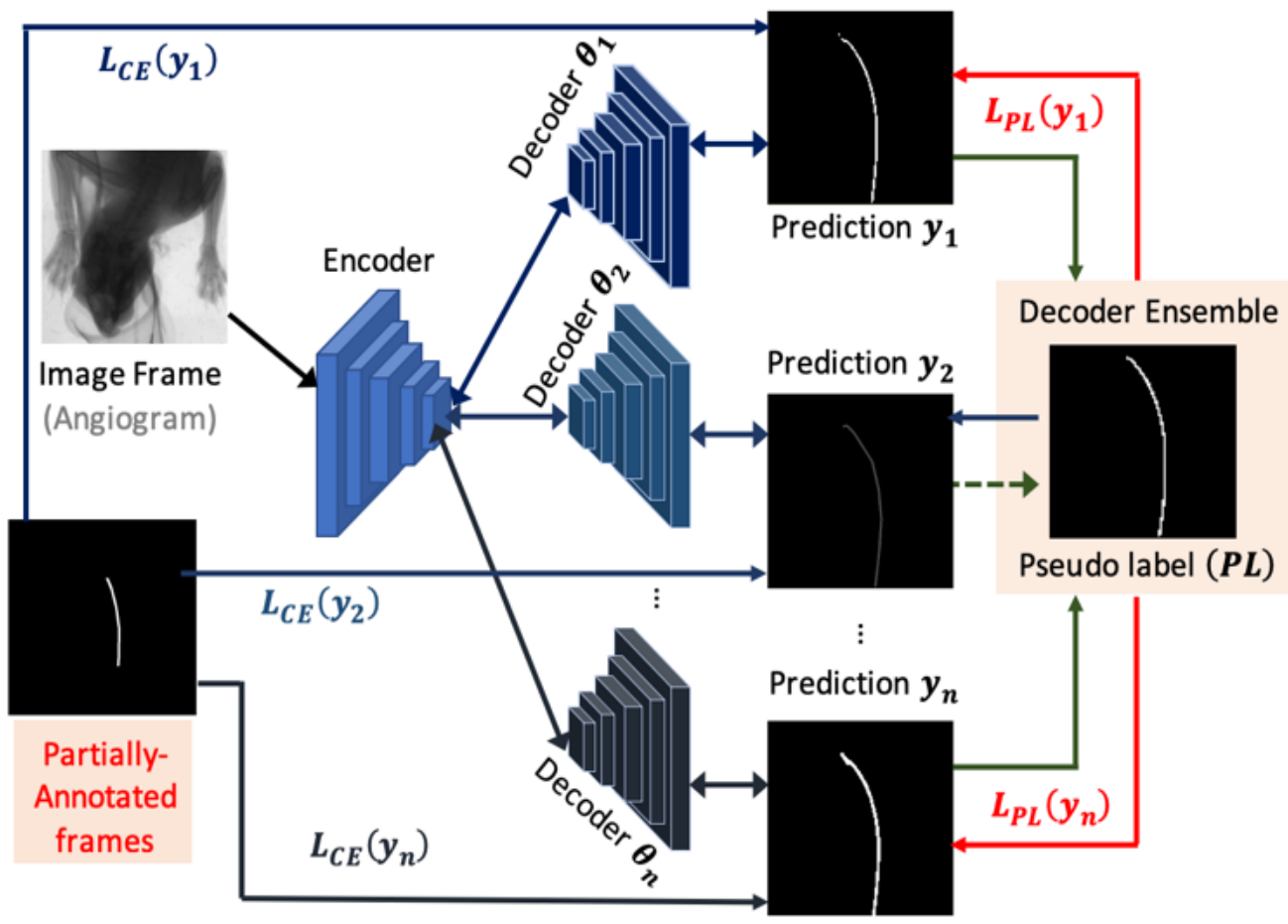

Fig. 1: Framework of weakly-supervised learning model with an encoder and laterally-branched multiple decoders.

Where $P(g_i = j|x_i) = \frac{\beta_i \mathcal{J}(x|\theta_i)}{\sum_{i=1}^{n} \beta_i \mathcal{J}(x|\theta_i)}$. The task is to use the limited annotated pixels to estimate the class distributions for the unlabeled components. Next, the estimated signals can be extended to enhance the model's performance beyond the supervised learning phase. For this, pseudo labels, derived as the mean prediction from one or more auxiliary decoders, are generated and combined with the outputs from a main decoder. Loss obtained from the pseudo labels and that from original annotations are jointly used to train the learning-based model.

### A. *Self-generating Pseudo Labels*

As depicted in Fig. 1, the weak-supervision model is based on an encoder-decoder structure, where one encoder feeds multiple branches of decoders. The encoder's feature maps are passed through the multiple laterally branched decoders, and their outputs are then combined to create the pseudo labels. The decoders differ by using dissimilar dilation rates to capture multi-scale feature maps. After initializing the training class distributions from the annotated signals, the outputs from the model's auxiliary decoders can be used to generate quasi labels by learning from the available supervision signals.

#### i. *Learning from Annotated Pixels*

The training dataset consists of angiogram images with partial annotations. This labeling approach includes pixels with sets of known and unknown labels. The annotated signals can be used to train a deep neural network by minimizing a loss function, assuming that the estimate of the ground-truth $(Y)$ can be provided by one-hot class values $\hat{Y} := \mathcal{H}(x|\Theta)$. During training, the network parameters $\Theta_{\mathcal{H}}$ are optimized by minimizing the loss using cross-entropy function $(\mathcal{L}_{CE})$, as defined in Eq. 3. Where $s$ is a one-hot annotation signal represented by the mixture component $g_i$ and pixel features $(x_i)$, while $y_i^c$ is the probability that $i^{th}$ pixel belongs to class $c$. Since the angiograms are partly annotated, pseudo labels are generated to simulate a fully-supervised learning experience.

$$\mathcal{L}_{CE}(y, s := g_i|x_i) = \sum_c \sum_{i=1}^{len(s)} log\ (y_i^c) \quad (3)$$

#### ii. *Generating Pseudo Labels*

The model uses multiple laterally branched decoders to propagate the available annotated pixels to the unlabeled ones. Each decoder utilizes the feature maps from the encoder to independently self-generate pseudo label signals. One decoder is designated as the main decoder and it is responsible for producing the actual pixel-level segmentation. The supervision signals from the main decoder are combined with the outputs of the auxiliary decoders to generate the pseudo labels. The probability maps are mixed as given in Eq. 4, where $\lambda$ are used to penalize the decoders' outputs. The $\lambda$ values are randomly chosen between 0 and 1, such that $\sum_{d=1}^{K} \lambda^d = 1$, per iteration.

$$PL(Mix_\lambda) = \underset{c}{\text{argmax}} \sum (\lambda^{d=1,\dots,K}) \times y_i^c \quad (4)$$

To increase the feature diversity, the outputs from the auxiliary decoders are perturbed during the training process [30, 31]. The perturbation implementation involves stochastic forward passes with random dropout. This process provides subsamples of the original model, denoted as $\mathcal{H}(y_i, \mathcal{J}(x|\Theta))$, where some of the features $\theta_i$ are randomly dropped. Next, the pseudo labels obtained from the unlabeled pixels are then combined with the supervised signals to augment the training.

### B. *Training with Shared Consistency*

By aggregating the outputs from multiple decoders, the model becomes more robust compared to using a single decoder. However, the random perturbations performed in the auxiliary decoders can introduce noise to the supervision signals obtained from the unlabeled pixels. This causes inconsistencies amongst the pseudo labels generated from the different decoders, causing variations in the sub-models' characteristics. Hence, consistency among the decoder outputs is required to filter the pseudo labels and enhance the model's learning experience. This is achieved by mixing the pseudo labels generated by the decoders with a shared consistency approach for training reliability [32]. Given a scenario with annotated labels $\mathcal{D}_L = \{x_i, y_i\}_{i=1}^{k}$ and pseudo labels $\mathcal{D}_p = \{x_i, y_i\}_{i=k+1}^{l}$, the overall training objective is to minimize the model's loss using both sets of labels, as derived in Eq. 5. Here, $y_{i,\dots,k}$ is the ground-truth and $y_{k+1,\dots,l}$ are the self-generated pseudo labels. In this setup, $l$ and $k$ are the number of labeled and unlabeled pixels, respectively. The objective function includes two parts: **supervised loss** that uses the ground-truth labels $y_{i,\dots,k}$ for minimizing the binary cross-entropy loss $(\mathcal{L}_{sup})$, and **pseudo-label loss** $(\mathcal{L}_{pse})$ which trains the model using the soft pseudo labels generated from Softmax learning combined with mean consistency loss function. The latter is designed to normalize the variations from auxiliary decoders.

$$\text{f}(x) = \min_{\theta} \left( \sum_{i=1}^{k} \mathcal{L}_{sup}(\mathcal{J}(x|\theta), y_i) + \sum_{i=1}^{l} \mathcal{L}_{pse}(\mathcal{J}(x;\ \theta, y_i)\ , \mathcal{J}(x; \theta', y_i')) \right) \quad (5)$$

The function includes a pairwise pseudo label supervision as further explained as expressed in Eqs. 6 - 7. For a given pair of decoders with pseudo labels: $PL_{d1}$ and $PL_{d2}$, the annotation signals from $PL_{d1}$ is used to supervise the signals from $PL_{d2}$, where $\mathfrak{D}(\cdot)$ is a distance measure between the class label distributions from the decoders. This offers shared consistency such that the different sub-models can co-learn, ensuring minimal variation amongst the decoders' outputs, thus improving the model's consistency. The final mixed loss function includes weighted combination of the segmentation loss added to the aggregated regularized loss to ensure the model's consistency, as derived in Eq. 7. The **consistency control parameter** ($\gamma$) is used to regulate the pseudo labels. With ablation study, we show that restricting the pseudo labels and using multiple decoders offers strong supervision signals and reduces mislabeling during training.

$$\mathcal{S}(PL_{d1}, PL_{d2}) = \frac{1}{K} \sum_{d,d' \in \mathcal{D}_u} \frac{1}{|\mathcal{D}_U|} \left( \sum \mathfrak{D}\big(g(\theta), g(\theta')\big) \right) \quad (6)$$

$$\mathcal{L}_{all} = \frac{1}{K} \left( \mathcal{L}_{pCE}(y_{d=1}, s) + \sum_{d' \forall d>1}^{K} \mathcal{L}_{pCE}(y_{d'}, s) \right) + \gamma \times \left( \sum_{\forall \{d1,d2\} in\ K}^{K} \mathcal{S}(PL_{d1}, PL_{d2}) \right) \quad (7)$$

## III. Implementation and Experiment Results

The proposed method was validated using U-Net [34]. The model's performance was evaluated based on the mIoU (mean intersection-over-union) values obtained for various datasets.

### A. Implementation Details

The segmentation model is implemented on a modified U-Net backbone [34]. The base model was extended to have three decoders with the first one serving as the main decoder. The other two are the auxiliary decoders used for generating pseudo labels from the partially-annotated data. The auxiliary decoders are replicas of the main decoder with addition of convolution and dropout layers. Thus, different feature maps are obtained from the three decoders, which help to prevent model overfitting. The model was implemented in Tensorflow Keras and validated on three different catheterization datasets. For training optimization, the image intensity was normalized, and data augmentation steps: zooming (0.2), translation (0.2), shearing (45°), rotation (45°), and flipping (0.5) were done.

The proposed weakly-supervised multi-lateral method was implemented with one encoder and three decoder branches. The model was trained and deployed for segmentation of tool pixels in the angiogram datasets. Each dataset was randomly partitioned into 90% for model training and 10% for model testing. The training and testing images were resized to 256 × 256 pixels for network input. Adam optimizer with an initial learning rate of $10^{-4}$ was used to minimize Eq. 3, where $\gamma$ was set as 0.25. The learning rate dynamically adjusted using drop and decay factors of 1 and 0.95, respectively, to improve the training convergence. Network weights were initialized with Xavier normal distribution and a mini-batch size of 4 was used as it gave the best validation. The model was trained on Nvidia A6000 RTX GPU for 200 epochs and the model with best performing weights was saved for evaluation and ablation studies, and application in a robotic catheterization setting.

### B. Experimental Results

Experiments were conducted to evaluate the proposed weakly-supervised method using three angiogram datasets.

#### i. Datasets and Preprocessing

***Dataset 1***: A private dataset consisting of angiograms obtained during a robotic catheterization study in rabbits. Ethics approval (SIAT-IRB-190215-H0291) was obtained, and the procedures were performed using the robotic catheter system (RCS) [33]. The angiograms, recorded intraoperatively, involved cannulating the rabbits' auricle-to-coronary vascular paths. A total of 2,700 angiograms (each 1440 × 1560 pixels with a resolution of 1.8 × 1.8 mm²) were preprocessed and used for offline model training. The data partially annotated in LabelMe [29] was used for end-to-end training and validation.

***Dataset 2***: This private dataset was obtained in robotic catheterization performed via the femoral-to-coronary artery of a pig (*weight:* 35.1 Kg). Institutional ethical approval (AAS-191204P) was obtained, and the procedure was carried out with our RCS platform. A total of 469 angiograms were recorded. The image frames have 768 × 768 pixels and were partially labeled with LabelMe for the model training purpose.

***Dataset 3***: A publicly available dataset of fluoroscopic images obtained from catheterization trials in a silicon aorta phantom. It consists of 2,000 angiograms extracted from four fluoroscopy videos [28] and has been validated for catheter segmentation in previous studies. The dataset is publicly available on https://weiss-develop.cs.ucl.ac.uk/fluoroscopy/, and has been validated for catheter segmentation in prior studies. Each image frame was resized to 256 × 256 pixels, and catheter pixels were manually annotated as full-scale binary masks, where background pixels have a value of "0" and catheter pixels are denoted by "1".

For this study, the annotations in all three datasets were adjusted by re-annotating 50% of the tool pixels (guidewire or catheter) to "0" (background pixels). These partially-annotated datasets were then used for end-to-end training and validation.

#### ii. Model Results

The segmentation results from the proposed model are presented in Fig. 2. Based on the confusion matrices on the left, the model achieved mIoU values of 70.81%, 67.06%, and 84.19% for the test sets in the rabbit, pig, and phantom data, respectively. These indicate the model's ability to effectively distinguish between tool and background pixels in the angiograms. Additional binary metrics such as precision, recall, sensitivity and specificity of the model are shown in the confusion matrix. For example, the model misclassified 0.07%, 0.25%, and 0.17% of background pixels as tool pixels in the rabbit, pig and phantom datasets, respectively. Receiver operating characteristic curves were also explored to analyze the model's aggregated performances for both classes. As shown on the right side of Fig. 2, the model achieved an average precision-recall with area under the curve of 74.54% 77.37%, 86.37% for test sets in the rabbit, pig, and phantom data, respectively. Thus, the model effectively separated the pixels to their classes of memberships with high probability.

## IV. Evaluation and Ablation Studies

### A. Evaluation Studies and Analysis

The segmentation performances of the proposed weakly- and fully-supervised models were compared. For the latter, the model was trained with the fully-annotated angiogram dataset. The segmentation outputs obtained by the models for the three datasets are as shown in Fig. 3 with no post-processing applied. These results include test image frames taken as model's input from each dataset, fully annotated data, and partially annotated data used as ground-truth, as shown in Fig. 3a-c, respectively. The white and black colored pixels in Figs. 3b and 3c are the actual tool and background pixels, respectively. The outputs from the proposed weakly-supervised model are displayed in Fig. 3d with the white and black colored pixels representing the segmentation outputs as tool and background pixels by the model, respectively. The proposed multi-lateral method, when implemented with three decoder branches achieved mIoU values of 70.81%, 67.06%, and 84.19% for the rabbit, pig, and phantom datasets, respectively. The fully-supervised model achieved higher mIoU values of 78.78%, 74.05%, and 90.42% for the same dataset (Fig. 4e). Thus, building the U-Net model with the proposed weakly-supervised method offers a closer performance with a mean margin of 9.24±1.5% compared to when the fully-supervised approach was used.

We also compared the proposed weakly-supervised multi-lateral decoder branching method with three existing weakly-supervised methods. These are the entropy minimization, total variation, and Mumford-Shah loss regularization [12]. The implementation of these methods follows the details in Sect. 3.1, and no post-segmentation procedures were applied. The segmentation results obtained with the three methods are also presented as shown in Figs. 4f-g. Using the fully-supervised method as the baseline, the percentage differences (mean±SD) observed are 16.40±4.99%, 27.64±8.18%, and 16.99±4.60% for entropy minimization, total variation, and Mumford-Shah loss regularization approaches, respectively. Therefore, the proposed method has the closest performance to the fully supervised method, and that is consistent across the datasets.

## B. *Ablation Studies*

### i. *Effect of multi-lateral branching*

We investigated the effects of using multiple decoders in the backbone. The U-Net architecture was redesigned into different versions having one decoder and two decoders. Each version was separately trained and evaluated on angiograms in ***Dataset 1***, and the results obtained from both versions were compared to the results from the three decoder configuration. The U-Net model with three decoders has an average mIOU of 69.22±0.74% (Fig. 4), which was higher than configurations with one decoder and two decoders. Using one decoder, which eliminates pseudo labeling, resulted in the lowest mIoU of 67.3±0.66%. This demonstrates that the single-decoder setup is less effective under weak supervision. The two decoder setup achieved an average mIoU of 68.41% ± 0.26%, similar to models using a single auxiliary decoder for pseudo labeling [12]. Therefore, mixing pseudo labels from multiple decoders enhances the segmentation performance.

### ii. *Effect of consistency thresholds*

We analyzed the effect of $\lambda$ threshold by using different values for the shared consistency loss in the three decoder setups. To ensure training stability and transparency, value of $\lambda^{d=1}$ *i.e.* the main decoder was fixed, while random values were used for $\lambda^{d>1}$. All $\lambda^{d=1}$ values were chosen between 0 and 1 with $\sum \lambda^{d\geq 1} = 1$ in each run. The models were trained for 100 epochs for each set of $\lambda$ values using the angiograms in ***Dataset 1***. We observed that the model's performance is sensitive to the $\lambda$ values as presented in Fig. 4. Although only slight variations were observed, the three-decoder setup gave a better performance by the systematic $\lambda$ selection. The optimal values of $\lambda$ for the models were 0.9, 0.3, and 0.5 for the one-decoder, two- decoder, and three-decoder setups, respectively.

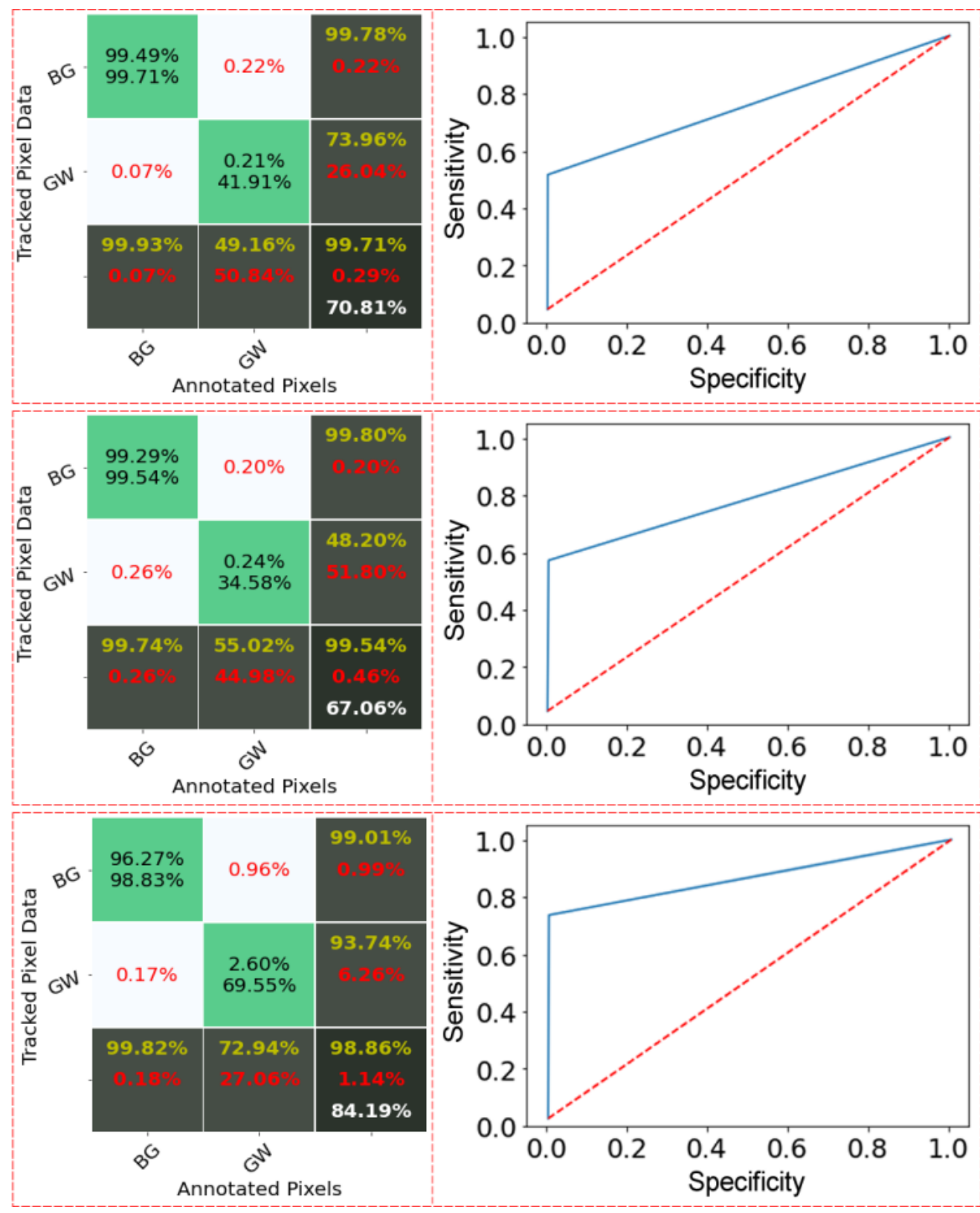

Fig. 2: Segmentation results for test sets in three angiogram datasets. (a) Rabbit data, (b) Pig data, and (c) Phantom data.

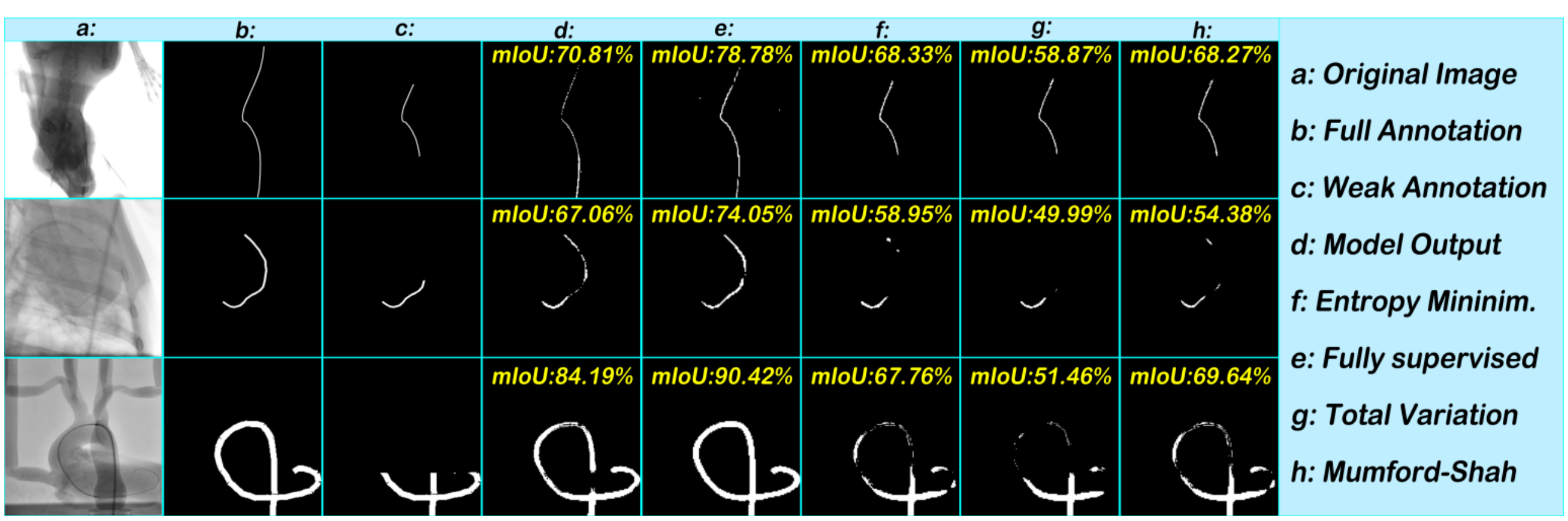

Fig. 3: Evaluation results for selected frames in the three test sets. *a)* original angiograms, *b)* fully annotated frames, and *c)* partial-annotated images. Segmentation outputs from the: *d)* proposed method, *e)* fully-supervised method, and *f-h)* existing weakly-supervised methods.

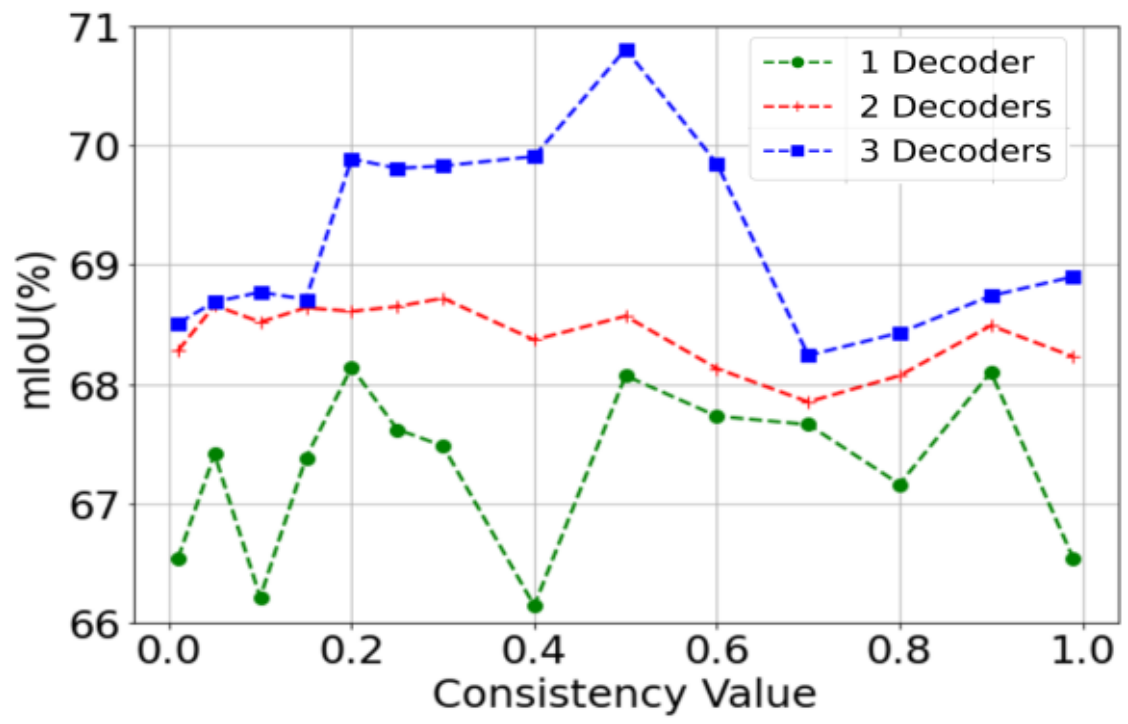

Fig. 4: Effects of different decoder numbers and consistency values

### *iii. Effects of loss functions on pseudo-label generation*

Pseudo-label generation relies on the features learnt from the partial annotations. Hence, we modified $\mathcal{L}_{CE}$ into a hybrid function and analyzed its contribution to the model. The new function is designed to down-weigh the pixels easy to classify and focus on the hard ones. The new function, in Eq. (7), uses two factors $\boldsymbol{\alpha}$ and $\boldsymbol{\beta}$ to regulate the learning process around the hard and soft pixels, where $\boldsymbol{X}$ and $\boldsymbol{Y}$ are true and predicted values, respectively. Additionally, we also removed the shared consistency to evaluate the effects of mixing loss function on the pseudo labeling. As in Table I, the results obtained show that the modified loss function and shared consistency offered 0.38% and 3.68% of the model's performance, respectively.

$$L^{XY} = \frac{1}{\beta(\alpha X + XY) + (1-\beta)\big(\alpha(1-X) + (1-X)(1-Y)\big)} \quad (7)$$

### *iv. Effect of backbone architecture*

We show that the proposed weakly-supervised method can work with other SOTA learning models. The multi-lateral model was implemented into four network architectures that are used for pixel-level object segmentation in medical image analysis. Each deep network was implemented with different percentages of annotated signals per image frame in ***Dataset-1***. The results obtained from each model are shown in Table II. The methods achieved an average mIoU of 58.95±5.55%, 66.64±5.0% and 72.42±4.27% when trained with 25%, 50%, and 75% partial annotation, respectively. It is seen that the five models achieved the closest performance to fully-supervised versions when trained with 75% partial annotation. Similarly, the performances by the proposed weakly-supervised method were stable across all the networks used in this study.

## V. APPLICATION IN ROBOT-ASSISTED CATHETERIZATION

The proposed weakly-supervised model was applied for online tool segmentation during robot-assisted catheterization using the leader-follower RCS reported in previous study [33]. As shown in Fig. 5a, the RCS was used to cannulate a silicon aorta phantom (Elastrat, Geneva, Switzerland) with a 0.014" guidewire and 6 fr catheter. Camera was used in place of angiography suite, and image frames (768 × 768 pixels) were extracted from video streams. The frames were preprocessed online, following the steps in Section 3, and sent as input to a pre-built weakly-supervised segmentation model. The model could offer online pixel-wise segmentation while navigating different vascular paths in the phantom with either guidewire or catheter. As presented in Fig. 5c, the proposed method gave high segmentation outputs for both the catheter and guidewire.

**Table I.** Performances in different pseudo-label generation modes

| Pseudo labeling mode | Performance (mIoU, %) | | |
|---|---|---|---|
| | *Dataset 1* | *Dataset 2* | *Dataset 3* |
| $\mathcal{L}_{sup}(\mathcal{I}(x\|\theta), y_i) = \mathcal{L}_{sup}^{XY}$ | 71.08 | 67.42 | 85.94 |
| $\mathcal{L}_{All} - \mathcal{S}(PL_{d1}, PL_{d2})$ | 68.56 | 63.85 | 81.63 |

**Table II.** Using SOTA models on different partial annotation sizes

| Models Used | Performance in rabbit data (%) | | | |
|---|---|---|---|---|
| | 25% | 50% | 75% | Full |
| **This Study** | 62.26 | 70.81 | 73.79 | 78.78 |
| Omisore *et al* [29] | **64.17** | **71.38** | **76.85** | **84.89** |
| Badrinarayanan [35] | 49.78 | 58.95 | 66.41 | 75.27 |
| Zhou *et al* [6] | 59.89 | 66.49 | 75.29 | 83.48 |
| Chen *et al* [36] | 58.63 | 65.57 | 69.76 | 76.24 |

**Table III.** Performances during robotic catheterization (Mean ± SD)

| Tool | mIoU (%) | Seg. Time (ms) | Connectivity (%) |
|---|---|---|---|
| **Guidewire** | 65.44±33.27 | 35.26±11.29 | 85.46±14.09 |
| **Catheter** | 86.83±12.79 | 33.85±13.64 | 93.94±15.24 |

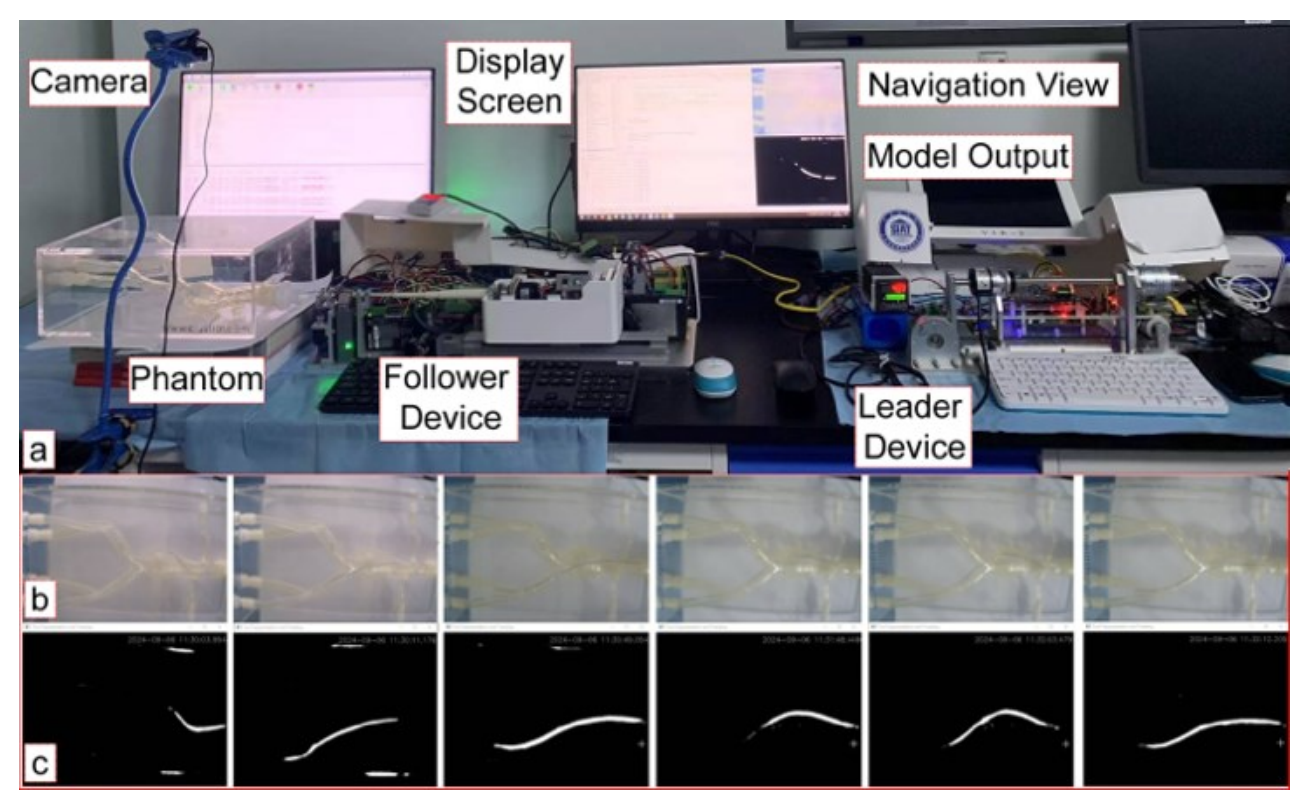

Fig. 5: Model deployment for tool visualization during robotic catheterization: a) Robot setup; b) Input image; c) Model output.

As shown in Table III, tool pixel tracking was observed with high segmentation performances (mIoU) of 65.44±33.27 and 86.83±12.79 % for guidewire and catheter. We analyzed the pixels' connectivity index of tools across all frames [19]. As in Table III, a high mean connectivity index was observed with mean segmentation time of ~35 ms per frame. This shows the percentage of continuous connected tool pixels intersecting with ground-truth pixels in the labels with 100% annotation.

## VI. CONCLUSION AND FUTURE WORKS

A weakly-supervised method with multi-lateral decoder branching is proposed for tool segmentation in catheterization data. The method is implemented in U-Net backbone and trained end-to-end with shared consistency loss for pixel-level pseudo labeling and segmentation. Supervision signals from three decoders are dynamically mixed for pseudo generation while the shared consistency function is used to enhance the model's performance. Experiments on three partial annotation catheterization data show the proposed method performs better than existing weakly-supervised methods, and closest to fully-supervised model. With end-point detection and pixel adjacent analytics [37], full connectivity can be obtained for whole tool segmentation during robot-assisted catheterization. This could aid automation of tasks like visual analytics and autonomous catheterization. Thus, the method be extended for real-time surgical scene analytics in robotic catheterization.

# REFERENCES

[1] Omisore O. M., et al., Towards Characterization and Adaptive Compensation of Backlash in a Novel Robotic Catheter System for Cardiovascular Interventions, IEEE Transactions on Biomedical Circuits and Systems, 2018, 12(4): p. 824-838.

[2] Naidu S. S., J. D. Abbott, J. Bagai, J. Blankenship, S. Garcia, S. N. Iqbal, P. Kaul, M. A. et al., SCAI expert consensus update on best practices in the cardiac catheterization laboratory, Catheterization and Cardiovascular Interventions, 2021, 98(2): p. 255-276.

[3] Chen C, Qin C, Qiu H, Tarroni G, Duan J, Bai W, Rueckert, Deep Learning for Cardiac Image Segmentation: A Review, Frontiers in Cardiovascular Medicine, 2020, 7(25).

[4] Baskaran L., G. Maliakal, S. J. Al'Aref, G. Singh, Z. Xu, K. Michalak, K. Dolan, U. Gianni, A. van Rosendael, I. van den Hoogen et al., Identification and Quantification of Cardiovascular Structures From CCTA: An End-to-End, Rapid, Pixel-Wise, Deep-Learning Method, JACC: Cardiovascular Imaging, 2020, 13(5): p. 1163-1171.

[5] Zhou Y. J., X. L. Xie, X. H. Zhou, S. Q. Liu, G. B. Bian, and Z. G. Hou, Pyramid attention recurrent networks for real-time guidewire segmentation and tracking in intraoperative X-ray fluoroscopy, Comput Med Imaging Graph, 2020, 83: p. 101734.

[6] Zhou Y. J., et al., A Real-Time Multifunctional Framework for Guidewire Morphological and Positional Analysis in Interventional X-Ray Fluoroscopy, IEEE Transactions on Cognitive and Developmental Systems, 2021, 13(3): p. 657-667.

[7] Ronneberger O., P. Fischer, and T. Brox, U-Net: Convolutional Networks for Biomedical Image Segmentation, in Medical Image Computing and Computer-Assisted Intervention – MICCAI 2015, Cham, 2015, 2015//, p.234-241.

[8] Zhou Z., M. M. Rahman Siddiquee, N. Tajbakhsh, and J. Liang, UNet++: A Nested U-Net Architecture for Medical Image Segmentation, in Deep Learning in Medical Image Analysis and Multimodal Learning for Clinical Decision Support, Cham, 2018, p.3-11.

[9] Gu Z., J. Cheng, H. Fu, K. Zhou, H. Hao, Y. Zhao, T. Zhang, S. Gao, and J. Liu, CE-Net: Context Encoder Network for 2D Medical Image Segmentation, IEEE Transactions on Medical Imaging, 2019, 38(10): p. 2281-2292.

[10] Dhamija T., A. Gupta, S. Gupta, Anjum, R. Katarya, and G. Singh, Semantic segmentation in medical images through transfused convolution and transformer networks, Applied Intelligence, 2022. 10.1007/s10489-022-03642-w.

[11] Calderon-Ramirez S. et al., Semisupervised Deep Learning for Image Classification With Distribution Mismatch: A Survey, IEEE Trans on Artificial Intelligence, 2022, 3(6): 1015.

[12] Luo X., M. Hu, W. Liao, S. Zhai, T. Song, G. Wang, and S. Zhang, Scribble-Supervised Medical Image Segmentation via Dual-Branch Network and Dynamically Mixed Pseudo Labels Supervision, MICCAI 2022, Cham, 2022, 2022//, p.528-538.

[13] Shin S. Y., S. Lee, I. D. Yun, S. M. Kim, and K. M. Lee, Joint Weakly and Semi-Supervised Deep Learning for Localization and Classification of Masses in Breast Ultrasound Images, IEEE Transactions on Medical Imaging, 2019, 38: p. 762-774.

[14] Wang J. and B. Xia, Bounding Box Tightness Prior for Weakly Supervised Image Segmentation, in MICCAI 2021, Cham, 2021, , p.526-536.

[15] Qu H. et al., Weakly Supervised Deep Nuclei Segmentation using Points Annotation in Histopathology Images, in MIDL2019

[16] Qian H. S., Li; Xuming, He, Weakly Supervised Volumetric Segmentation via Self-taught Shape Denoising Model, in Machine Learning Research2021, p.268–285.

[17] Viniavskyi O., M. Dobko, and O. Dobosevych, Weakly-Supervised Segmentation for Disease Localization in Chest X-Ray Images, ArXiv, 2020, abs/2007.00748.

[18] Ibrahim M. S., A. A. Badr, M. R. Abdallah, I. F. Eissa, Bounding Box Object Localization Based On Image Superpixelization, Procedia Computer Science, 2012, 13: p. 108-119.

[19] Omisore O. M., et al., Automatic tool segmentation and tracking during robotic intravascular catheterization for cardiac interventions, Quantitative imaging in medicine and surgery, 2021, 11 6: p. 2688-2710.

[20] Girum K. B., et al., Fast interactive medical image segmentation with weakly supervised deep learning method, Int J Comput Assist Radiol Surg, 2020, 15(9): p. 1437-1444.

[21] Rajchl M., et al., DeepCut: Object Segmentation From Bounding Box Annotations Using Convolutional Neural Networks, IEEE Trans Med Imaging, 2017, 36(2): p. 674-683.

[22] Valvano G., A. Leo, and S. A. Tsaftaris, Learning to Segment From Scribbles Using Multi-Scale Adversarial Attention Gates, IEEE TMI, 2021, 40(8): p. 1990-2001.

[23] Zhang K., X. Zhuang, ShapePU: A New PU Learning Framework Regularized by Global Consistency for Scribble Supervised Cardiac Segmentation, in MICCAI 2022, p.162-172.

[24] Zhang K. Z., Xiahai, CycleMix: A Holistic Strategy for Medical Image Segmentation From Scribble Supervision, IEEE/CVF CVPR 2022, p.11656-11665.

[25] Shuwei Z. G., Wang; Xiangde, Luo; Qiang, Yue; Kang, Li; Shaoting, Zhang. (2022). PA-Seg: Learning from Point Annotations for 3D Medical Image Segmentation using Contextual Regularization and Cross Knowledge Distillation. .

[26] Issam L., et al., A Weakly supervised consistency-based learning method for {covid-19} segmentation in CT Images, in IEEE Winter Conference on Applications of Computer Vision, Waikoloa, HI, USA, 2021, January 3-8, 2021, p. 2452-2461.

[27] Yang H., C. Shan, A. F. Kolen, P. H. Weakly-supervised learning for catheter segmentation in 3D frustum ultrasound, Comp Med Imag. Graph., 2022, 96: p. 102037.

[28] Gherardini M., E. Mazomenos, A. Menciassi, and D. Stoyanov, Catheter segmentation in X-ray fluoroscopy using synthetic data and transfer learning with light U-nets, Computer Methods and Programs in Biomedicine, 2020, 192: p. 105420.

[29] O.M. Omisore, T. Akinyemi, W. Duan, W. Du, and L. Wang, "Multi-lateral Branched Network for Endovascular Tool Segmentation during Robot-assisted Catheterization", IEEE Transactions in Medical Robotics and Bionics. 6(2): 433-447, Jan. 2024.

[30] Ouali Y., C. Hudelot, and M. Tami, Semi-Supervised Semantic Segmentation With Cross-Consistency Training, in 2020 IEEE/CVF CVPR, 13-19 June 2020, p.12671-12681.

[31] Wu Y., M. Xu, Z. Ge, J. Cai, and L. Zhang, Semi-supervised Left Atrium Segmentation with Mutual Consistency Training, in MICCAI 2021, Cham, 2021, p.297-306.

[32] Verma V., Kawaguchi, A. Lamb, J. Kannala, A. Solin, Y. Bengio, D. Lopez, Interpolation consistency training for semi-supervised learning, Neural Networks, 2022, 145: p. 90-106.

[33] W. Duan, L. Zihao, O.M. Omisore, W. Du, T. Akinyemi, X.Y Chen, X. Gao, H. Wang, and L. Wang, "Development of an Intuitive Interface with Haptic Enhancement for Robot-Assisted Endovascular Intervention," IEEE Transactions on Haptics, PP(99):1-13. Dec. 2023, https://doi.org/10.1109/TOH.2023.3346479.

[34] Zheng Y., M. J. Er, S. Shen, W. Li, Y. Li, W. Du, W. Duan, and O. M. Omisore, An Improved Image Segmentation Model based on U-Net for Interventional Intravascular Robots, 4th International Conference on Intelligent Autonomous Systems, 2021, p.84-90.

[35] Badrinarayanan V., A. Kendall, and R. Cipolla, SegNet: A Deep Convolutional Encoder-Decoder Architecture for Image Segmentation, IEEE Transactions on Pattern Analysis and Machine Intelligence, 2017, 39(12): p. 2481-2495.

[36] Chen L.-C., Y. Zhu, G. Papandreou, F. Schroff, and H. Adam, Encoder-Decoder with Atrous Separable Convolution for Semantic Image Segmentation, in ECCV 2018, p.833.

[37] W. Du, G. Yi, O.M. Omisore, W. Duan, X. Chen, T. Akinyemi, J. Liu, B. G. Lee, and Lei Wang, Guidewire Endpoint Detection Based on Pixel-Adjacent Relation during Robot-Assisted Intravascular Catheterization: In Vivo Mammalian Models, Advanced Intelligent Systems, 2300687, pp. 1-17, 2023.